\pdfoutput=1

\documentclass[11pt]{article}

\usepackage[]{latex/acl}

\usepackage{times}
\usepackage{latexsym}
\usepackage{booktabs}
\usepackage{scalerel}

\usepackage[T1]{fontenc}

\usepackage[utf8]{inputenc}

\usepackage{microtype}

\usepackage{inconsolata}
\usepackage{graphicx}
\usepackage{amsmath, amssymb, amsfonts}
\usepackage{bbm}
\usepackage{multirow}
\usepackage{float}

\def\vA{{\mathbf{A}}}
\def\cd{{\mathcal{D}}}

\def\va{{\mathbf{a}}}

\def\vk{{\mathbf{k}}}

\def\vt{{\mathbf{t}}}
\def\vx{{\mathbf{x}}}

\usepackage{inconsolata}

\title{Evaluating and Improving Continual Learning in Spoken \\Language Understanding}

\author{
Muqiao Yang$^1$\quad Xiang Li$^1$\quad Umberto Cappellazzo$^2$\\ \textbf{Shinji Watanabe$^1$\quad Bhiksha Raj$^{1,3}$}\\
$^1$ Carnegie Mellon University\quad $^2$ University of Trento\\ $^3$ Mohamed bin Zayed University of Artificial Intelligence\\
\texttt{\{muqiaoy, xl6, bhiksha\}@cs.cmu.edu,} \\ \texttt{umberto.cappellazzo@unitn.it, shinjiw@ieee.org}}

\begin{document}
\maketitle

\begin{abstract}
Continual learning has emerged as an increasingly important challenge across various tasks, including Spoken Language Understanding (SLU). In SLU, its objective is to effectively handle the emergence of new concepts and evolving environments. The evaluation of continual learning algorithms typically involves assessing the model's stability, plasticity, and generalizability as fundamental aspects of standards. However, existing continual learning metrics primarily focus on only one or two of the properties. They neglect the overall performance across all tasks, and do not adequately disentangle the plasticity versus stability/generalizability trade-offs within the model. In this work, we propose an evaluation methodology that provides a unified evaluation on stability, plasticity, and generalizability in continual learning. By employing the proposed metric, we demonstrate how introducing various knowledge distillations can improve different aspects of these three properties of the SLU model. We further show that our proposed metric is more sensitive in capturing the impact of task ordering in continual learning, making it better suited for practical use-case scenarios.
\end{abstract}

\section{Introduction}
\vspace{-3mm}

Spoken Language Understanding (SLU) focuses on interpreting and understanding human speech in order to extract meaningful information from it \cite{tur2011spoken}. With the ubiquity of voice assistants and home devices, SLU systems have become an integral part of our society. While traditional SLU systems make use of the concatenation of an automatic speech recognition (ASR) and a natural language understanding block \cite{mesnil2014using, coucke2018snips}, recently, end-to-end (E2E) strategies have gained traction, buttress-
\begin{figure}[H]
    \centering
    \includegraphics[scale=0.31]{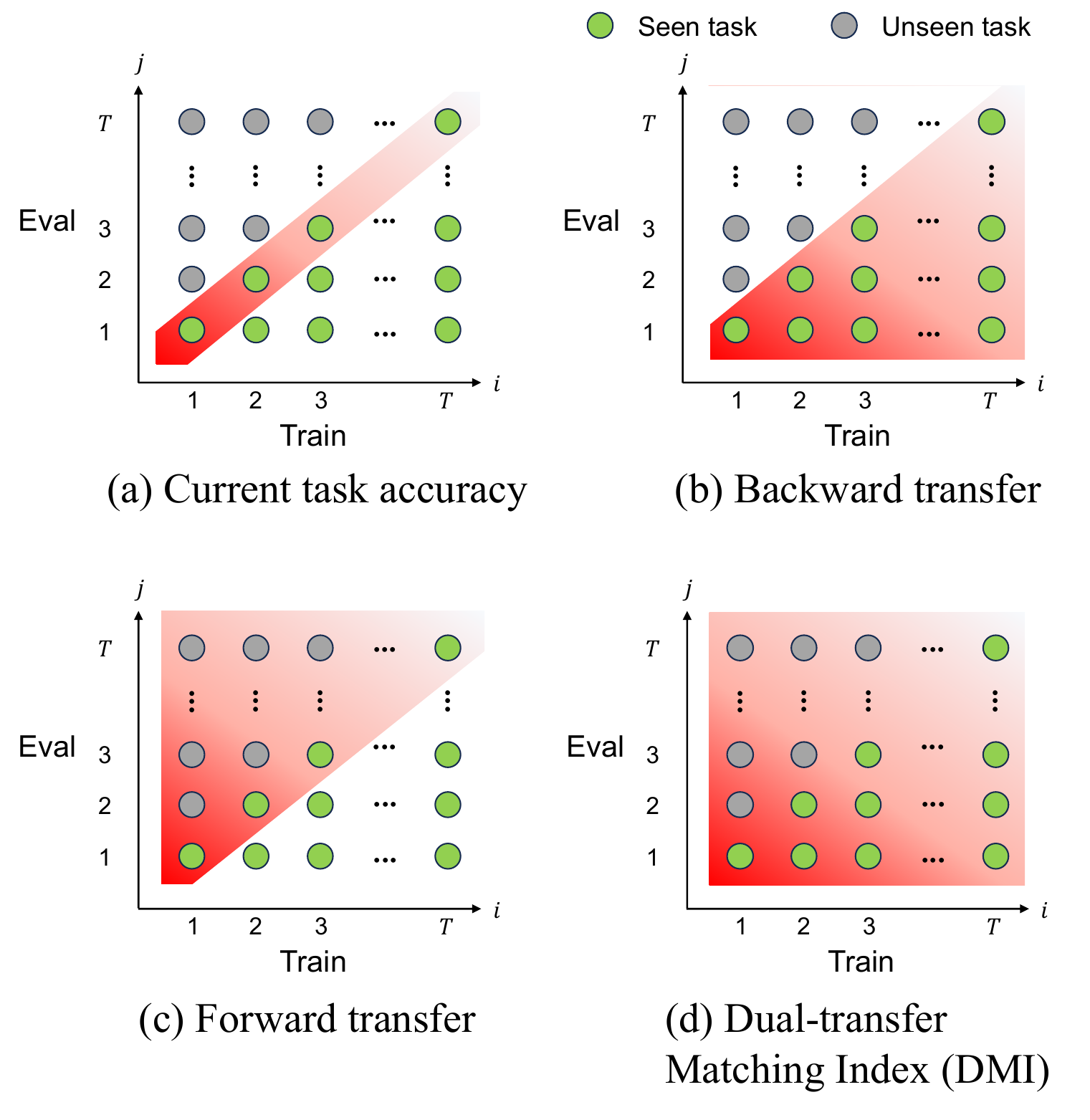}
    \caption{An illustration of the proposed Dual-transfer Matching Index (DMI) vs. other continual learning related metrics, including backward transfer and forward transfer \cite{lopez2017gradient}. The vertical and horizontal axes represent the sequence of tasks presented to the network for learning. $T$ is the total number of tasks. A circle at index $(i,j)$ means the evaluation on task $j$ after finishing training task $i$. A green circle indicates the evaluated performance of one seen task, while a grey circle is assessing unseen classes in future tasks. By covering the whole $T \times T$ matrix, our DMI provides an evaluation on three aspects of model capabilities, including stability, plasticity and generalizability.}
    \label{fig:teaser}
\end{figure}
\noindent ed by their reduced latency and error propagation \cite{saxon2021end, arora2022token}. 



SLU systems are often expected to learn to recognize a continually expanding ``vocabulary'' of intents, as they encounter them in training data -- a task known as continual or lifelong learning. Without appropriate techniques, unlike human cognition, SLU models are susceptible to a phenomenon called catastrophic forgetting  \cite{french1999catastrophic}, where previously learned knowledge is lost when acquiring new information. Overlooking previously acquired knowledge or skills as newer ones are acquired can hinder the performance of the models. Therefore, continual learning aims to circumvent the forgetting effect with the goal of continually adapting the model by learning from infinite streams of data \cite{kirkpatrick2017overcoming}. An ideal SLU model should be \textit{stable}, retaining previously acquired knowledge, while also being \textit{plastic} enough to  learn new intents effectively. Furthermore, it should also be \textit{generalizable}, retaining the ability to learn novel yet-to-be-encountered information. Evaluating the effectiveness of continual learning algorithms requires measuring their performance across these three criteria.




However, existing continual learning evaluation metrics have certain limitations. Among stability, plasticity, and generalizability, most mainstream continual learning evaluation metrics focus on only one or two aspects of the three properties. A brief illustration is provided as a train-evaluation matrix in Figure \ref{fig:teaser}. By immediately evaluating the performance after learning one task, current task accuracy (Figure \ref{fig:teaser}a) only focuses on the diagonal entries of the matrix, thus measuring the plasticity of the model. Backward transfer and forward transfer (Figure \ref{fig:teaser}b and \ref{fig:teaser}c) evaluate the stability and generalizability by focusing on the lower and upper triangular matrix \cite{lopez2017gradient}. However, they only measure the partial impact of the current task on either past or future tasks, neglecting the overall performance of all seen and unseen tasks. Meanwhile, these metrics do not fully disentangle the plasticity and stability/generalizability trade-offs in continual learning, as their computation involves operations on the diagonal entries in the matrix. As a result, they may not fully capture the dynamic behaviors of models during the training process, and are less sensitive to the order of tasks particularly when the contents of adjacent tasks exhibit large semantic variations.

In this paper, we propose Dual-transfer Matching Index (DMI) as a unified continual learning metric to evaluate all three properties: stability, plasticity and generalizability. We compute the DMI by employing the Hungarian maximum matching algorithm to compare model predictions with ground-truth intent labels for all seen and unseen classes at the end of each task. This approach allows us to disentangle and quantify the three properties in a comprehensive manner. Based on that evaluation using DMI, we apply various knowledge distillation (KD) techniques to enhance the three properties respectively. We demonstrate that introducing different levels of KDs can improve all three standards and provide a better understanding of the continual learning behaviors of SLU models. Furthermore, we show that our proposed DMI metric is more sensitive than previous metrics in capturing the effect of task ordering.
 
\section{Evaluation in Continual Learning}


\subsection{Stability-plasticity Dilemma}
\label{subsec: dilemma}

\begin{figure}
    \centering
    \includegraphics[scale=0.4]{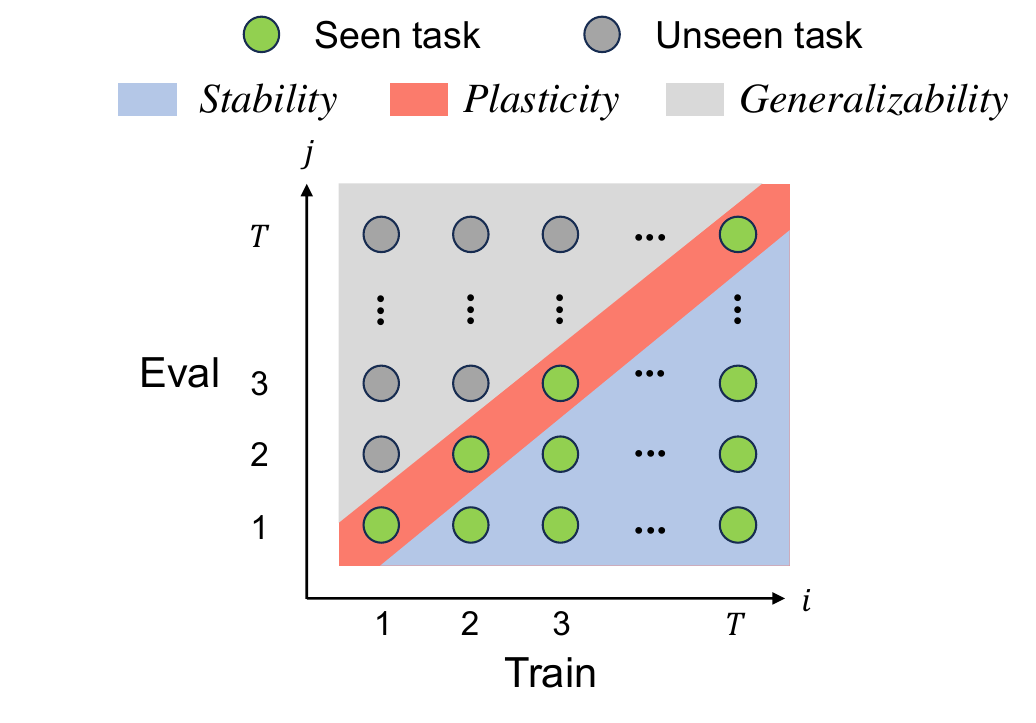}
    \caption{Train-evaluation performance matrix $\mathbf{A}$ in continual learning. The diagonal entries (red) represent the plasticity with the performance of current tasks.  The lower triangular matrix (blue) represents the focused field of stability, while the upper triangular matrix (grey) measures generalizability to unseen tasks.}
    \label{fig:s_p_g}
\end{figure}

The main issue that continual learning aims to address was formulated as the stability-plasticity dilemma \cite{mermillod2013stability}. An ideal system is expected to possess both plasticity, enabling the rapid acquisition of new knowledge, and stability, preventing the forgetting effect of previously learned knowledge.

As the two sides of the same coin, stability and plasticity are often regarded as a trade-off between learning and remembering. Assume that a model is continuously encountering data from different tasks $\{\cd_1, \cd_2, \cdots, \cd_T\}$ in chronological order, where $T$ indicates the total number of tasks. Among different $\cd_t$s, the distribution of data and labels may shift across tasks. If the model has a high memory stability, it will remain stable and not easily erase the knowledge learned in the past tasks, thus may hinder the capability of the model to swiftly accommodate new data distributions. Conversely, excessive plasticity may impede the stability of the system by prioritizing adapting to distribution changes. Therefore, different metrics are often applied to assess the stability and plasticity of continual learning algorithms.

We formulate the metric evaluation in continual learning with the train-evaluation performance matrix $\mathbf{A}$, which is a $T \times T$ matrix where $T$ is the total number of tasks. Each entry $\mathbf{A}_{i, j}$ represents the evaluated performance on task $j$ after the model has learned task $i$. Figure \ref{fig:s_p_g} provides a visual representation of the matrix. The entries can be classified into two categories, where the green circles represent the seen tasks and the grey circles represent the unseen tasks.

In this work, we establish the taxonomy of evaluation metrics based on the property that a metric is trying to assess in the model, namely stability, plasticity, and generalizability. We propose a novel metric and compare it with existing metrics in terms of formulation and experimental results. By considering different perspectives, it provides valuable insights into the evaluation of continual learning systems.


\subsection{Formulation of Continual Metrics}
\label{subsec: prior}

Based on the train-evaluation performance matrix, we provide a formulation of the existing continual learning metrics. Along the diagonal of the train-evaluation matrix as in Figure \ref{fig:s_p_g}, the circles indicate the immediate evaluation of task $i$ after learning the same task. Therefore, it reflects the plasticity of the model by showing how well and fast the model is adapting to new knowledge. Current task accuracy (cur-ACC) is one common metric to assess the plasticity by calculating the average accuracies of tasks from $1$ to current task $t$ along the diagonal of the evaluation matrix:
\begin{equation}
\vspace{-2.5mm}
    \text{cur-ACC}_t = \frac{1}{t} \sum_{i=1}^t \mathbf{A}_{i,i}  , t\in\{1,\cdots, T\}
\end{equation}

The lower triangular matrix of $\mathbf{A}$ reflects the stability. At task $t$, it covers the evaluation results of all previous tasks $1$ to $t-1$. One of the most straightforward stability-based metrics is last accuracy (last-ACC), which assesses the stability by calculating the average accuracy of all tasks after the last task $T$:
\begin{equation}
\vspace{-2.5mm}
    \text{last-ACC} = \frac{1}{T} \sum_{i=1}^T \mathbf{A}_{T, i}
\end{equation}

However, last-ACC provides only one-time evaluation at the end and does not cover all entries in the lower triangular matrix. Therefore, it provides little information on the stability of the entire learning process. To better quantify the overall stability, backward transfer (BWT) measures the performance degradation of an arbitrary task $t$ on its previous tasks from $1$ to $t-1$ \cite{lopez2017gradient}. It calculates the average differences between the evaluation on a previous task $j$ after learning current task $i$ ($\mathbf{A}_{i, j}$), and the immediate evaluation of the previous task $j$ ($\mathbf{A}_{j,j}$), i.e.,

\vspace{-4mm}
\begin{equation}
\begin{aligned}
\text{BWT}_t = \frac{2}{t(t-1)} \sum_{i=2}^t\sum_{j=1}^{i-1}(\mathbf{A}_{i, j} - \mathbf{A}_{j, j}) , \\
t\in\{2,\cdots, T\}
\end{aligned}
\end{equation}

Typically, the BWT score is a negative number, indicating that losing past information is almost inevitable during the learning process of new tasks. 

Forward transfer (FWT) is a metric used to assess generalizability. It is mostly used as an opposite metric compared to BWT. We note that we refer to the implementation of the Continuum library \footnote{https://github.com/Continvvm/continuum}, which evaluates the influence that learning task $t-1$ has on the performance of the future incoming task $t$. Since the model has not seen task $t$ yet, FWT quantifies the generalizability by calculating the gain of the current model performance on the incoming future task ($\mathbf{A}_{i-1,i}$), compared with zero-shot evaluation results on the prospective task ($\bar{\mathbf{A}}_i$), i.e.,

\vspace{-4mm}
\begin{equation}
    \text{FWT}_t = \frac{1}{t-1} \sum_{i=2}^{t} (\mathbf{A}_{i-1, i} - \bar{\mathbf{A}}_{i}), t\in\{2,\cdots, T\}
\end{equation}

\section{Continual Learning with Unified Evaluation}

\subsection{DMI: Dual-transfer Matching Index}

In this work, beyond limited and entangled evaluations of stability, plasticity, or generalizability, we propose an evaluation metric named Dual-transfer Matching Index (DMI), to provide a unified evaluation for continual learning. Unlike existing metrics discussed in Section \ref{subsec: prior} that only show how the knowledge transfers in a unidirectional manner, either backward or forward, our metric evaluate the continual learning behavior of the model in terms of all three properties. Meanwhile, instead of entangling plasticity versus stability and generalizability, our DMI provides a disentangled quantification for each of the three properties.

Assume that our training process elapses along the x-axis of the train-evaluation performance matrix $\mathbf{A}$. The model is continuously learning on a sequence of tasks $1, 2, \dots, T$. At the end of training task $i$, instead of evaluating the accuracy of one single previous or future task, we evaluate the current model on all seen and unseen classes. However, since the model has not yet gathered information for tasks $t \succ i$, we can only evaluate these tasks in a class-agnostic manner. 

Let us denote the total number of data samples as $N$ and the number of classes as $K$. Specifically, we first extract the predicted intent class embeddings $\{\vx^{(i)}_n\}_{n=1}^N$ at each task $i$. Then we perform k-means clustering $\mathcal{K}(\cdot)$ \cite{macqueen1967classification, lloyd1982least} on $\vx^{(i)}$, with the total number of clusters as $K$. By denoting $\boldsymbol{\mu}_k^{(i)}$ as the centroid of cluster $k$ at task $i$, we can obtain the class-agnostic assignment as 

\begin{equation}
\begin{aligned}
    & \vk^{(i)} = \mathcal{K}(\vx^{(i)}), \\
    & k_n^{(i)} = \arg\min_k |\vx_n^{(i)} - \boldsymbol{\mu}_k^{(i)}|
\end{aligned}
\end{equation}

At each task $i$, $\vk^{(i)}=(k_1^{(i)}, k_2^{(i)}, \dots, k_N^{(i)})$ indicates the class-agnostic assignment for all $N$ samples. Class-agnostic means that $\vk^{(i)}$ is not aware of the exact class index, as there exist unseen tasks that the model has not learned from yet. Then we compute the matching score between the class-agnostic assignment $\vk^{(i)}$ and the ground-truth class labels $\vk_{\text{gt}}$ using Hungarian maximum matching algorithm $\mathcal{H}(\cdot, \cdot)$ \cite{kuhn1955hungarian}. As a result, $\mathcal{H}(\vk^{(i)}, \vk_{\text{gt}})$ gives us an optimal matched mapping from $\vk^{(i)}$ to $\vk_{\text{gt}}$. After that, we calculate the accuracies of the best mapping with the ground-truth labels to populate the train-evaluation performance matrix $\vA$, i.e.,

\begin{figure}
    \centering
    \includegraphics[scale=0.53]{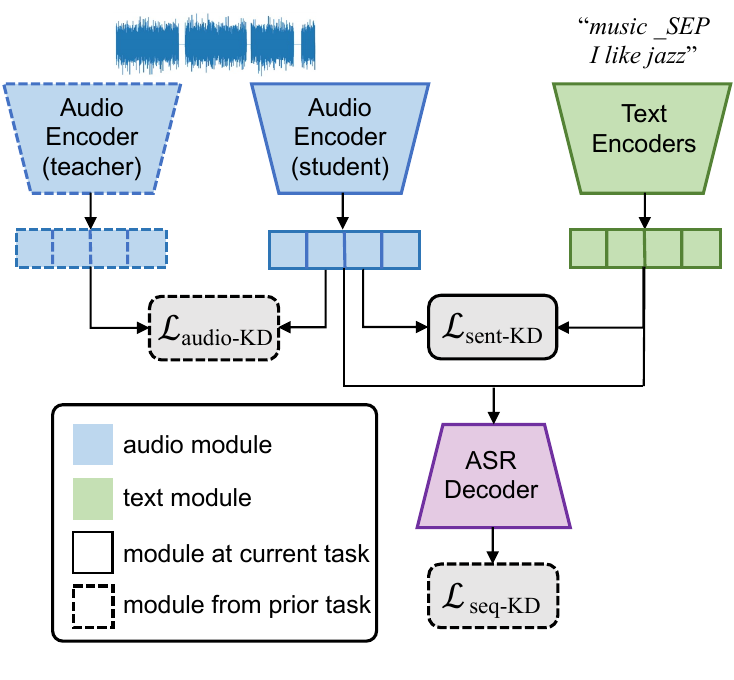}
    \caption{Pipeline overview of our SLU training. Dashed blocks indicate the knowledge distillation from previous tasks.}
    \label{fig:pipeline}
\end{figure}

\begin{equation}
\begin{aligned}
    & \vk^{(i)*} = \mathcal{H}(\vk^{(i)},\vk_{\text{gt}}),  \\
    & \vA_{i,j} = \frac{\sum_{k\in\mathcal{J}} \sum_{n} \mathbbm{1}(k_n^{(i)*} \text{==\ } k_{\text{gt}, n} \text{==\ } k)}{\sum_{k\in\mathcal{J}} \sum_n\mathbbm{1}(k_{\text{gt,n}} \text{==\ }k)}
\end{aligned}
\end{equation}

\noindent where $\mathbbm{1}(\cdot)$ returns $1$ if the condition is true and $0$ otherwise. $k_n^{(i)*}$ and $k_{\text{gt}, n}$ indicate the $n$th element of $\vk^{(i)*}$ and $\vk_{\text{gt}}$ respectively. $\mathcal{J}$ refers to the set of classes that appear at task $j$. The numerator calculates the total number of correctly matched samples that belong to task $j$. The denominator is the total number of samples in task $j$. Therefore, $\vA_{i,j}$ is represented as the class-agnostic accuracy of task $j$ after training on task $i$.

We perform similar computations for each task $i$ from $1$ to $T$ to fulfill all entries of the matrix $\vA$. Finally, we compute the DMI as a quantification for stability, plasticity, and generalizability respectively by taking the mean of the entries in the corresponding regions as shown in Figure \ref{fig:s_p_g}:

\begin{equation}
\begin{aligned}
    & \text{DMI}_{stab} = \frac{2}{T(T-1)} \sum_{j=1}^{i-1}\sum_{i=1}^T \vA_{i,j}, \\
    & \text{DMI}_{plas} = \frac{1}{T} \sum_{i=1}^T \vA_{i,i}, \\
    & \text{DMI}_{gen} = \frac{2}{T(T-1)} \sum_{i=1}^{j-1}\sum_{j=1}^T \vA_{i,j} 
\end{aligned}
\end{equation}

Note that we complete the performance matrix $\vA$ with prior metrics in different ways. We compute the entries of matrix $\vA$ by the overall clustering accuracies of both past and future tasks, as well as the current task. In this way, DMI consists of three separate scores to measure the stability, plasticity, and generalizability during the training process of the model. Therefore, it is expected to provide a disentangled quantification of the three properties, and reflect how the learned knowledge accumulates and evolves in a unified manner.

\subsection{Pipeline Formulation}
\label{subsec: pipeline}

We establish our pipeline in a class-incremental learning (CIL) setting in SLU. Specifically, the model is continuously adapted to a sequence of different tasks, and incremental intent labels emerge sequentially across tasks. CIL is a challenging setting in real-world scenarios, as the model is agnostic to task labels during inference time \cite{hsu2018re}.

We utilize a combined architecture for automatic speech recognition (ASR) and SLU, as it is shown that the auxiliary use of the ASR transcriptions can lead to better SLU performances \cite{arora2022token, cha2021speak}. Under this setting, we prepend intent tokens to the original transcription and separate them using a special token $\langle \textit{\_SEP} \rangle$. This operation extends the transcription and transforms the modeling into a sequence-to-sequence (seq2seq) problem. As a result, our architecture diverges from the conventional continual learning pipelines that consist of a feature extractor followed by a classifier. Instead, we employ a single ASR decoder that takes the encoder outputs and generates predicted tokens in a sequential manner. Assume that the input audio-text pair is denoted as $(\va, \vt)$. The goal of the ASR decoder is to find the most probable extended transcription sequence $\vt$ given the audio input $\va$, i.e.,

\vspace{-3mm}
\begin{equation}
\begin{aligned}
    & \arg\max_{\vt} P (\vt | \va, \theta) \\
& = \arg\max_{\vt} P (\vt_1, \vt_2, \cdots, \vt_M | \va, \theta) \\
& = \arg\max_{\vt}\prod_{m=2}^{M} P(\vt_m|\vt_{m-1}, \va, \theta)
\end{aligned}
\end{equation}

\noindent where $\theta$ is the parameter set of the model and $M$ is the length of the transcription. Conditional independence is assumed at the last step. $\vt_2$ is the intent token, succeeding $\langle \textit{CLS} \rangle$ as the first token. The formulation of the overall pipeline is illustrated in Figure \ref{fig:pipeline}.

\subsection{Knowledge Distillations}
\label{subsec: kd}

The objective of knowledge distillation (KD) \cite{hinton2015distilling} aligns with continual learning by transferring knowledge from a teacher model to a student model. In continual learning, the teacher could refer to the model learned from previous tasks or a pretrained large network, while the student means the model that is learning currently. It is noteworthy that a small portion of previous data (typically smaller than $5\%$), together with its extracted features, is saved into one rehearsal memory for the future use of teacher models. We apply multiple types of KDs to improve the stability, plasticity, and generalizability of the SLU model. Figure \ref{fig:pipeline} shows the multiple KD techniques that we have leveraged in our pipeline. 

\paragraph{Audio-KD.} Catastrophic forgetting at the audio encoder is one of the main factors that affect the continual learning performance of SLU. To address this issue, we introduce KD at the output of the audio encoder. Let us denote the teacher audio encoder and student audio encoder as $h_{i-1}^a$ and $h_i^a$ respectively. $i-1$ and $i$ are task indices, meaning that we would like to distill the knowledge from previously learned models to the current student model. Assume that the sampled rehearsal memory is $\mathcal{M}$. We add an audio-KD loss to main the continuity between the audio representations produced by the student audio encoder at current task $i$ and the teacher at previous task $i-1$ for the rehearsal data, i.e., 

\vspace{-3mm}
\begin{equation}
    \mathcal{L}_{\mathrm{audio-KD}}=\frac{1}{|\mathcal{M}|}\sum_{(\va, \vt)\in\mathcal{M}} \langle h_{i-1}^a(\va), h_i^a(\va) \rangle
\end{equation}

\noindent where $\langle \cdot, \cdot \rangle$ indicates cosine similarity. The objective of audio-KD is to regularize the change of the parameters of the audio encoder, such that the student audio encoder retains partial information from the teacher audio encoder. Therefore, it is expected to improve the stability of the model.

\paragraph{Seq-KD.} As we have discussed in Section \ref{subsec: pipeline}, our pipeline employs an encoder-decoder architecture to predict the intents. Similar to the audio encoder, the ASR decoder may also suffer from catastrophic forgetting. To address this issue, we also introduce a sequence KD at the output of ASR decoder to reduce the forgetting effect at a sequence level. It works by pushing the student model to generate sequences that are close to the sequence-level distribution of the teacher model. The seq-KD loss is defined as 

\vspace{-3mm}
\begin{equation}
\begin{aligned}
    & \mathcal{L}_{\mathrm{seq-KD}} =-\frac{1}{|\mathcal{M}|}\sum_{\scaleto{(\va, \hat{\vt})\in\mathcal{M}}{7.5pt}}\log P(\hat{\vt}|\va, \theta) \\
    &= -\frac{1}{|\mathcal{M}|}\sum_{\scaleto{(\va, \hat{\vt})\in\mathcal{M}}{7.5pt}}\log \prod_{m=2}^M P(\hat{\vt}_m| \hat{\vt}_{m-1}, \va, \theta) \\
\end{aligned}
\end{equation}

\noindent where $\theta$ is the parameter set of the student model. $\hat{\vt}$ is the output sequence generated with beam search using the teacher model, and it is saved into the rehearsal memory $\mathcal{M}$ together with the paired audio $\va$ during the previous task. 

\paragraph{Sent-KD.} Both audio-KD and seq-KD aim to increase the stability of the system by distilling knowledge from previous tasks. They are operated at the encoder and decoder levels of the pipeline respectively. In addition to them, we introduce another KD from a pretrained text encoder to our student audio encoder. Its objective is to increase the plasticity and generalizability of the model with pretrained knowledge. With distilled knowledge from the pretrained text representations, the learned audio representation can be aligned to the shared embedding space. By doing so, the generalizability of the model are expected to be improved, as the learned audio embedding is assumed to contain the similar semantic information as the text input in the sentence level. 

Let us denote the pretrained text encoder as $h^t$. It remains frozen during our training process, while the audio encoder is fine-tuned. Since the output of the audio encoder is a sequential embedding, we need to first average-pool the audio embedding vectors to reduce their temporal dimension, so that they are comparable with the sentence embedding. Then, the sentence KD is defined as 
\begin{equation}
\begin{aligned}
    & \bar{h}_i^a(\va) = \mathrm{pool}(h_i^a(\va)), \\
    & \mathcal{L}_\mathrm{sent-KD}=||h^t(\vt) - \bar{h}_i^a(\va)||^2
\end{aligned}
\end{equation}

And the final loss function to update the model is summarized as 
\begin{equation}
    \begin{aligned}
        \mathcal{L} &= \mathcal{L}_{\mathrm{CE}} + \lambda_\mathrm{audio} \mathcal{L}_{\mathrm{audio-KD}}  \\
        & \quad + \lambda_\mathrm{seq} \mathcal{L}_{\mathrm{seq-KD}} +\lambda_\mathrm{sent}\mathcal{L}_\mathrm{sent-KD}
    \end{aligned}
\end{equation}

\noindent where $\mathcal{L}_\mathrm{CE}$ refers to the cross entropy loss between the predicted and ground-truth token logits. $\lambda_\mathrm{audio}$, $\lambda_\mathrm{seq}$ and $\lambda_\mathrm{sent}$ are coefficients to control the weight of corresponding KDs.

Overall, we introduce a combination of different knowledge distillation techniques to boost the stability, plasticity, and generalizability of our continually learning SLU model. In the next section, we will empirically show the improvement of the quantification of different metrics under various settings.
\section{Experiments}

\subsection{Experimental Setup}
\label{subsec:exp_setup}

\begin{table*}[]
    \centering
    \begin{tabular}{l|c|c|c|c|c|c|c|c}
    \toprule
    \toprule
       Dataset & \multicolumn{4}{c|}{FSC} & \multicolumn{4}{c}{SLURP} \\
       Metric  & Acc & BWT & FWT & DMI & Acc & BWT & FWT & DMI \\
    \midrule
       Fine-tune  & 29.9  & -61.2  & 0 & 22.1 / 63.6 / 25.7  & 31.9 & -64.5 & 0 & 20.2 / 61.7 / 22.4 \\
       Random  & 68.6 & -16.3 & 63.2 & 63.7 / 78.2 / 24.9 & 66.6 & -18.7 & 63.5 & 63.8 / 75.3 / 22.7   \\
       \quad + audio-KD & 72.9  & -13.3 & 67.3 & 73.0 / 75.9 / 25.6 & 68.1 & -15.9 & 63.9 &  70.3 / 70.8 / 24.7  \\
       \quad + seq-KD & 75.1 & -10.2 & 71.4 & 76.4 / 75.6 / 25.3 & 68.9 & -14.7 & 65.0 &  72.5 / 70.9 / 25.8  \\
       \quad + sent-KD & 76.7 & -10.1 & 72.9 & 76.9 / 75.7 / 28.8 & 70.3 & -14.0 & 66.8 & 73.2 / 71.3 / 28.6   \\
       Herding  & 69.8 & -15.4 & 66.3  & 66.3 / \textbf{78.4} / 24.8 & 68.1 & -16.2 & 63.7 & 64.5 / \textbf{75.8} / 23.1   \\
       \quad + audio-KD & 73.5 & -11.7 & 70.1 & 78.3 / 75.7 / 25.9 & 68.7 & -12.8 & 64.0 &  74.4 / 72.9 / 24.7  \\
       \quad + seq-KD & 75.7 & -9.6 & 72.9 & 79.1 / 75.1 / 26.4 & 70.6 & -12.3 & 67.1 & 76.9 / 72.3 / 25.2   \\
       \quad + sent-KD & \textbf{77.9} & \textbf{-9.0} & \textbf{73.3} & \textbf{80.0} / 75.7 / \textbf{30.8} & \textbf{71.1} & \textbf{-12.1} & \textbf{67.9} & \textbf{77.8} / 70.5 / \textbf{30.4}   \\
    \bottomrule
    \bottomrule
    \end{tabular}
    \caption{Quantitative evaluation results in terms of Acc ($\uparrow$), BWT ($\uparrow$), FWT ($\uparrow$), and our proposed DMI ($\uparrow$). The three numbers of DMI are in the order of stability, plasticity, and generalizability.}
    \label{tab:exp}
\end{table*}

We base our experiments on the intent classification of Fluent Speech Commands (FSC) \cite{lugosch2019speech} and Spoken Language Understanding Resource Package (SLURP) \cite{bastianelli2020slurp} datasets. FSC includes 30,043 English utterances, recorded at 16 kHz. It provides 248 different utterances that are mapped in 31 different intents (i.e., our classes). SLURP  contains roughly 72K audio recordings of single-turn user interactions with a home assistant, annotated with scenarios, actions and entities. Overall, there are 18 unique scenarios and 69 intents. Following \cite{cappellazzo2023sequence}, we divide the SLURP dataset into tasks using the scenario labels as splitting criterion, whereas for FSC we use the intents. We choose to divide the datasets into $T=6$ tasks, with each task containing different classes. Implementation details can be found in Appendix \ref{app: impl}.

By default, we establish the order of tasks based on the principle of prioritizing the most populated scenarios. This is expected to ensure the model to learn richer information in early tasks, thus mitigating the catastrophic forgetting effect. We will provide the overall evaluation results with the default setting in Section \ref{subsec: exp}, and further investigate the effect of task orders in the later Section \ref{subsec: order}.


\subsection{Evaluation Results and Analysis}
\label{subsec: exp}

We evaluate the continual learning performance on both FSC and SLURP with metrics introduced in Section \ref{subsec: prior} including accuracy (Acc), backward transfer (BWT), forward transfer (FWT) \cite{lopez2017gradient} and our proposed DMI. The accuracy is reported as last-ACC in Section \ref{subsec: prior}. The experimental results are shown in Table \ref{tab:exp}.

The first row in the table is the naive fine-tuning result, without any continual learning algorithms applied. Therefore, it serves as the lower bound of our performance. The accuracy is relatively low and the BWT gives a substantial negative number, which offers us a general understanding of the extent to which the forgetting effect impacts the performance in the absence of continual learning techniques.

Recall that we mentioned in Section \ref{subsec: kd} that we are saving a portion of past data as the rehearsal memory for the future use of later tasks. In the implementation, we set the ratio as $1\%$. To retrieve samples from the rehearsal memory, we employ two sampling strategies: random sampling and herding-based sampling \cite{rebuffi2017icarl}. Different from random sampling, herding-based sampling strategy selects the samples that are closest to a small subset of exemplars in their class. This is expected to make the sampling process more robust to the changes in data representations, thus improving the continual learning performance.

From Table \ref{tab:exp}, the results of random and herding-based sampling both yield significantly better performance than the fine-tuning experiment, demonstrating improved continual learning performance. We can also observe that herding-based sampling performs almost consistently better than random sampling. 

Additionally, we incorporate various KD techniques as described in Section \ref{subsec: kd} to validate their effectiveness. As we have introduced previously, audio-KD and seq-KD are expected to increase the stability of the system by regularizing parameter changes at the encoder and decoder levels. By leveraging semantic information from a pretrained text encoder, sent-KD may increase the generalizability of continual learning. Table \ref{tab:exp} presents the results of adding three KDs sequentially. We can observe that both BWT and the first number of DMI ($\text{DMI}_{stab}$) are increased after adding audio-KD and seq-KD, indicating that the stability of the model is improved. On the other hand, sent-KD boosts the generalizability by leveraging pretrained semantic information, reflected in the increase of FWT and the third number of DMI ($\text{DMI}_{gen}$). However, as the effect of stability-plasticity dilemma discussed in Section \ref{subsec: dilemma}, the plasticity of the model ($\text{DMI}_{plas}$) might be decreased as a trade-off when KD techniques are added to regularize the model. The reason is that we are penalizing the parameter changes in the model with KDs from the teacher model of the previous task, thus potentially slowing down the learning process on the current task. Such a phenomenon is not reflected in prior metrics.

We also note that both BWT and FWT improve with the introduction of additional KD techniques. This is due to their entanglement with the plasticity as shown in Figure \ref{fig:teaser}, therefore may not fully reflect the stability/generalizability of the model. As a comparison, our DMI metric provides a disentangled and unified evaluation of all three properties, and a more comprehensive view of how the model behavior changes and evolves during the continual learning process.

\begin{figure}
    \centering
    \includegraphics[scale=0.25]{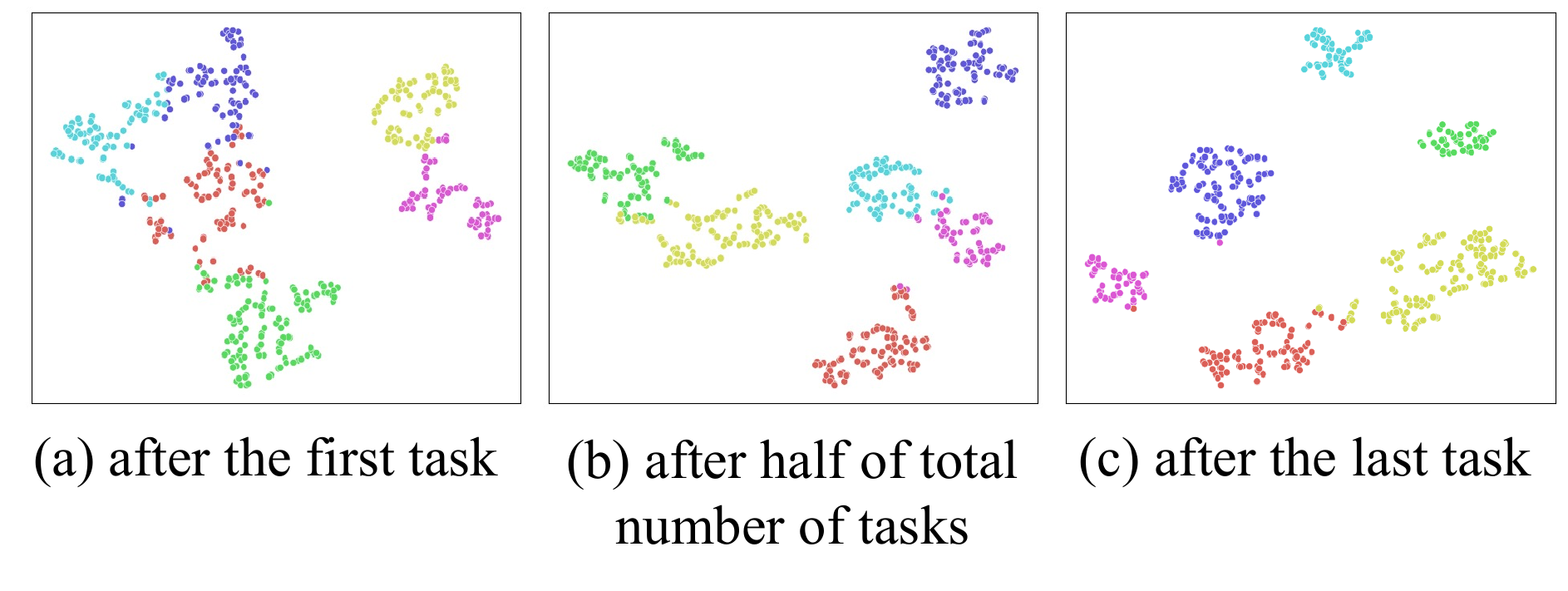}
    \caption{Qualitative results to show the change of clustering across tasks.}
    \label{fig:vis}
\end{figure}

In addition to quantitative performance, we show the qualitative results of clustering with t-SNE \cite{van2008visualizing} in Figure \ref{fig:vis}. For visualization purposes, we select one class from each task to plot. Figure \ref{fig:vis}(a) to (c) depict how the class-agnostic clustering evolves during the learning process from the first task to the last. The greater the distance between clusters of different colors, the better the model's performance. In the beginning, although the model has not encountered 5 out of the 6 classes from later tasks yet, it still exhibits some generalizability to distinguish unseen classes from seen intents. As the continual learning process progresses, the model starts to gain increasingly better capability to classify intents from each other. Finally, Figure \ref{fig:vis}(c) provides a clustering result with each of the classes clearly separated. This validates the increasing effectiveness and generalizability of our continual learning SLU model from a qualitative perspective.

\subsection{Effect of Task Ordering on Evaluation}
\label{subsec: order}

\begin{table}[]
    \centering
    \scalebox{0.92}{
    \begin{tabular}{c|c|c|c}
    \toprule
      & \multirow{2}{*}{\shortstack{\small Freqency\\ \small Order}} & \multirow{2}{*}{\shortstack{\small Close-semantic\\ \small Order}} & \multirow{2}{*}{\shortstack{\small Diverse-semantic \\ \small Order}} \\
      & & & \\
    \midrule
     \small  Acc & 77.9 & 75.6 & 73.9 \\
     \small  BWT & -9.0 & -9.1 & -10.9 \\
     \small  FWT & 73.3 & 74.9 & 69.3  \\
      $\small \text{DMI}_{stab}$ & 80.0 & 71.5 & 60.2  \\
      $\small \text{DMI}_{plas}$ & 75.7 & 60.2 & 58.2  \\
      $\small \text{DMI}_{gen}$ & 30.8 & 44.7 & 50.4  \\
    \bottomrule
    \end{tabular}}
    \caption{Evaluation results for the effect of different task orderings on metrics.}
    \label{tab:order}
\end{table}

Continual learning performance is affected by the ordering of tasks \cite{bell2022effect}. In the case of SLU, since each transcription and its intent have different semantic meanings, the learning order of tasks may impact the overall performance, including stability, plasticity, and generalizability.

To validate this, we name the ordering scheme introduced in Section \ref{subsec:exp_setup} as \textit{frequency order}, as it is ranking tasks in terms of their frequencies. Alternatively, we choose to rank the tasks based on their semantic meanings in two opposite ways, namely \textit{close-semantic order} and \textit{diverse-semantic order}. Close-semantic order groups together data instances with similar semantic meanings within the same task. For example, utterances with intents ``bring newspaper'' and ``bring shoes'' are grouped together to establish a close semantic relationship within one task. Under the setting of diverse-semantic order, the intents are strategically arranged to ensure a mix of semantic meanings in close proximity.

The effect of different tasks orderings on metrics is presented in Table \ref{tab:order}. By using the DMI metric, we observe that both close-semantic and diverse-semantic orders improve the generalizability while decreasing the stability and plasticity compared to the frequency order. The impact of diverse-semantic order is particularly more significant. The reason behind this may be that frequency order benefits stability and plasticity by exposing the model to a larger amount of data, while close-semantic order assists the model by grouping semantically similar classes together within a task. On the other hand, diverse-semantic order consists of a wider range of classes in each task, thereby equipping the model with a higher generalizability. However, as a trade-off, the complexity of individual tasks increases, resulting in decreased stability and plasticity of the model.

As a comparison, these changing tendencies are not effectively captured by prior metrics. From Table \ref{tab:order}, BWT and FWT are mostly correlated with Acc, and a higher Acc would mostly lead to higher BWT and FWT scores. This is because these metrics are entangled with plasticity, and may not fully reflect the stability and generalizability. In contrast, our DMI metric provides a more sensitive measurement of stability, plasticity, and generalizability when the order of tasks changes. This makes our metric better suited for practical use-case scenarios.


\section{Related Work}


\vspace{-2mm}
While the majority of research on continual learning has focused on computer vision, there have been notable efforts to extend it to speech and text domains. Regarding speech, it has been explored for the problems of Keyword Spotting \cite{xiao2022rainbow}, ASR \cite{yang2022online, chang2021towards, diwan2023continual}, and SLU \cite{cappellazzo2022exploring, cappellazzo2023sequence}. For NLP, some works aim to equip language models with continual learning capabilities. For example, \cite{razdaibiedina2023progressive} propose a prompt learning-based method, whereas \cite{ke2023continual} combines soft masking units and contrastive learning to alleviate forgetting. An investigation of the taxonomy of these continual learning-related approaches is introduced in Appendix \ref{app: cl}. However, there has been limited prior research on comprehensive continual evaluation metrics. Although \cite{de2022continual} proposes a new set of metrics to quantify the worst-case performance in previous tasks, it differs from our proposed metric as it does not provide a unified evaluation encompassing multiple aspects of evaluations in the continual learning system.

\section{Conclusion}

\vspace{-2mm}
In this paper, we propose Dual-transfer Matching Index as one new evaluation metric for continual learning. It provides a unified and disentangled evaluation in terms of stability, plasticity, and generalizability. Moreover, we show that introducing multiple knowledge distillation techniques helps the SLU model improve all three standards. We also empirically demonstrate that our metric is more sensitive than previous evaluation metrics in capturing the effect of class ordering.
\section{Limitations}

In this work, we experiment with spoken language understanding as the main target task by applying the proposed evaluation metric. However, the concept could also be adapted to other similar classification tasks. Specifically in SLU, our contributions include both the proposed generalized DMI evaluation metric and the improved SLU training with different knowledge distillation methods to reflect the improvement of stability, plasticity, and generalizability measured by DMI. In other relevant tasks, due to different model architectures and task settings, the observations of the evaluated stability/plasticity/generalizability will also be likely to be different, and specific techniques to improve the three aspects of capabilities in continual learning models might be dependent on the selected task.

\bibliography{custom}

\appendix

\section{Appendix}
\label{sec:appendix}

\subsection{Implementation Details}
\label{app: impl}

For both datasets, the text encoder utilizes a standard text embedding layer of size 768. We use Sentence-BERT \cite{reimers2019sentence} for the pretrained text encoder to extract the sentence-level information. Regarding the audio encoder, we employ a pre-trained and fine-tuned base Wav2vec 2.0 model \cite{baevski2020wav2vec} on 960 hours of Librispeech for SLURP ($\sim$94.3M parameters), while for FSC, we utilize the base DistilHuBERT \cite{chang2022distilhubert} ($\sim$23.5M parameters). As FSC is a less challenging dataset compared to SLURP, we discovered that a smaller pre-trained encoder is sufficient to achieve state-of-the-art results. Both encoders have a hidden size of 768, and their feature extractor remains fixed during training. Similar to \cite{radford2021learning}, we employ linear projection layers to map the representations from each encoder to the audio-text embedding space, which has a dimension of 512. The ASR decoder is based on the transformer model with 6 layers, a hidden size of 768, 8 attention heads, and feedforward layers with a dimension of 2048.

For tokenization, we utilize Byte-Pair Encoding (BPE) \cite{sennrich2015neural} with a vocabulary size of 1,000 and a BPE dropout rate of 0.1 for SLURP. However, for FSC, due to the limited number of unique words, we use word tokenization resulting in 139 tokens. BPE automatically assigns a dedicated token to each intent, while for FSC, we manually add the intent tokens. During inference and for the computation of soft labels for seq-KD, we apply beam search with a beam width of 10 for FSC and 20 for SLURP, respectively. We use the validation set to tune the hyperparameters and select the best model for each task. All the weights of KDs are used as $0.1$.

\subsection{Taxonomy of continual learning approaches} 
\label{app: cl}

Following the standard nomenclature, continual learning strategies can be grouped into a few categories, according to the approach they rely on \cite{wang2023comprehensive, zhou2023deep}. \textit{Regularization}-based methods introduce ad-hoc regularization terms to combat forgetting. Some methods regularize the weights of the network \cite{kirkpatrick2017overcoming, schwarz2018progress, aljundi2018memory, park2019continual, wang2021afec, adel2019continual}, whereas others penalize changes to the model's intermediate or final outputs, usually by means of the knowledge distillation (KD) concept \cite{hinton2015distilling}. By denoting the model trained in the previous task as the teacher model and that trained in the current task as the student model, KD fosters the pass of the knowledge accrued in the teacher model onto the student. We can identify \textit{logit} distillation methods \cite{li2017learning, rebuffi2017icarl} that align the predictions of the student and teacher models, and \textit{feature} distillation that works on the intermediate feature embeddings produced by the feature encoders \cite{douillard2020podnet, hu2021distilling}. It is also possible to combine logit and feature distillations together \cite{cappellazzo2022exploring}.

\textit{Replay}-based methods either maintain a memory buffer where a bunch of old training samples are collected \cite{rolnick2019experience, bang2021rainbow} (experience replay) or generate data samples by means of a dedicated generative model \cite{shin2017continual, xiang2019incremental} (generative replay). Finally, \textit{architecture}-based approaches introduce task-specific parameters, either by expanding the network itself \cite{yan2021dynamically, zhou2022model} or keeping the network frozen and learning a small amount of additional parameters (i.e., prompts) \cite{wang2022learning, smith2023coda}. 

\end{document}